\PassOptionsToPackage{table}{xcolor}
\documentclass[sigconf]{acmart}
\usepackage{bbding,balance,hyperref,multirow}

\AtBeginDocument{%
  \providecommand\BibTeX{{%
    \normalfont B\kern-0.5em{\scshape i\kern-0.25em b}\kern-0.8em\TeX}}}

\copyrightyear{2023}
\acmYear{2023} 
\setcopyright{rightsretained} 
\acmConference[MM '23]{Proceedings of the 31st ACM International Conference on Multimedia}{October 29-November 3, 2023}{Ottawa, ON, Canada}
\acmBooktitle{Proceedings of the 31st ACM International Conference on Multimedia (MM '23), October 29-November 3, 2023, Ottawa, ON, Canada}
\acmDOI{10.1145/3581783.3611753}
\acmISBN{979-8-4007-0108-5/23/10}

\makeatletter
\gdef\@copyrightpermission{
\begin{minipage}{0.3\columnwidth}
\href{https://creativecommons.org/licenses/by/4.0/}{\includegraphics[width=0.90\textwidth]{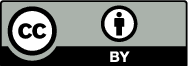}}
\end{minipage}\hfill
\begin{minipage}{0.7\columnwidth}
\href{https://creativecommons.org/licenses/by/4.0/}{This work is licensed under a Creative Commons Attribution International 4.0 License.}
\end{minipage}
\vspace{5pt}
}
\makeatother

\begin{document}

\title{Separate and Locate: Rethink the Text in Text-based Visual Question Answering }

\author{Chengyang Fang}

\email{fangchengyang@iie.ac.cn}
\affiliation{%
  \institution{Institute of Information Engineering, Chinese Academy of Sciences}
  \city{ }
  \country{ }
}
\affiliation{%
  \institution{School of Cyber Security, University\\  of Chinese Academy of Sciences}
  \city{ }
  \country{ }
}

\author{Jiangnan Li}

\email{lijiangnan@iie.ac.cn}
\affiliation{%
  \institution{Institute of Information Engineering,\\ Chinese Academy of Sciences}
  \city{ }
  \country{ }
}
\affiliation{%
  \institution{School of Cyber Security, University\\  of Chinese Academy of Sciences}
  \city{ }
  \country{ }
}

\author{Liang Li}
 \authornote{Corresponding author }

\email{liliang@iie.ac.cn}
\affiliation{%
  \institution{Institute of Information Engineering,\\ Chinese Academy of Sciences}
  \city{ }
  \country{ }
}
\affiliation{%
  \institution{School of Cyber Security, University \\of  Chinese Academy of Sciences}
  \city{ }
  \country{ }
}

\author{Can Ma}

\email{macan@iie.ac.cn}
\affiliation{%
  \institution{Institute of Information Engineering,\\ Chinese Academy of Sciences}
  \city{ }
  \country{ }
}
\affiliation{%
  \institution{School of Cyber Security, University\\  of Chinese Academy of Sciences}
  \city{ }
  \country{ }
}

\author{Dayong Hu}

\email{superhudayong@163.com}
\affiliation{%
  \institution{Heilongjiang Network Space Research
Center}
  \city{ }
  \country{ }
}


\begin{abstract}
  Text-based Visual Question Answering (TextVQA) aims at answering questions about the text in images.
  Most works in this field focus on designing network structures or pre-training tasks.
  All these methods list the OCR texts in reading order (from left to right and top to bottom) to form a sequence, which is treated as a natural language ``sentence''. However, they ignore the fact that most OCR words in the TextVQA task \textbf{do not have a semantical contextual relationship}. 
  In addition, these approaches use 1-D position embedding to construct the spatial relation between OCR tokens sequentially, which is not reasonable. The 1-D position embedding \textbf{can only represent the left-right sequence relationship between words in a sentence}, but not the complex spatial position relationship.
  To tackle these problems, we propose a novel method named Separate and Locate (SaL) that explores text contextual cues and designs spatial position embedding  to construct spatial relations between OCR texts. 
  Specifically, we propose a Text Semantic Separate (TSS) module that  helps the model recognize whether words have semantic contextual relations. 
  Then, we introduce a Spatial Circle Position (SCP) module that helps the model better construct and reason the spatial position relationships between OCR texts.
Our SaL model outperforms the baseline model by 4.44\% and 3.96\% accuracy on TextVQA and ST-VQA datasets. Compared with the pre-training state-of-the-art method pre-trained on 64 million pre-training samples, our method, without any pre-training tasks, still achieves 2.68\% and 2.52\% accuracy improvement on TextVQA and ST-VQA. Our code and models will be released at {\href{https://github.com/fangbufang/SaL }{https://github.com/fangbufang/SaL} }.
\end{abstract}




\maketitle

\begin{figure*}[!htb]
  \centering
  \includegraphics[width=0.9\linewidth]{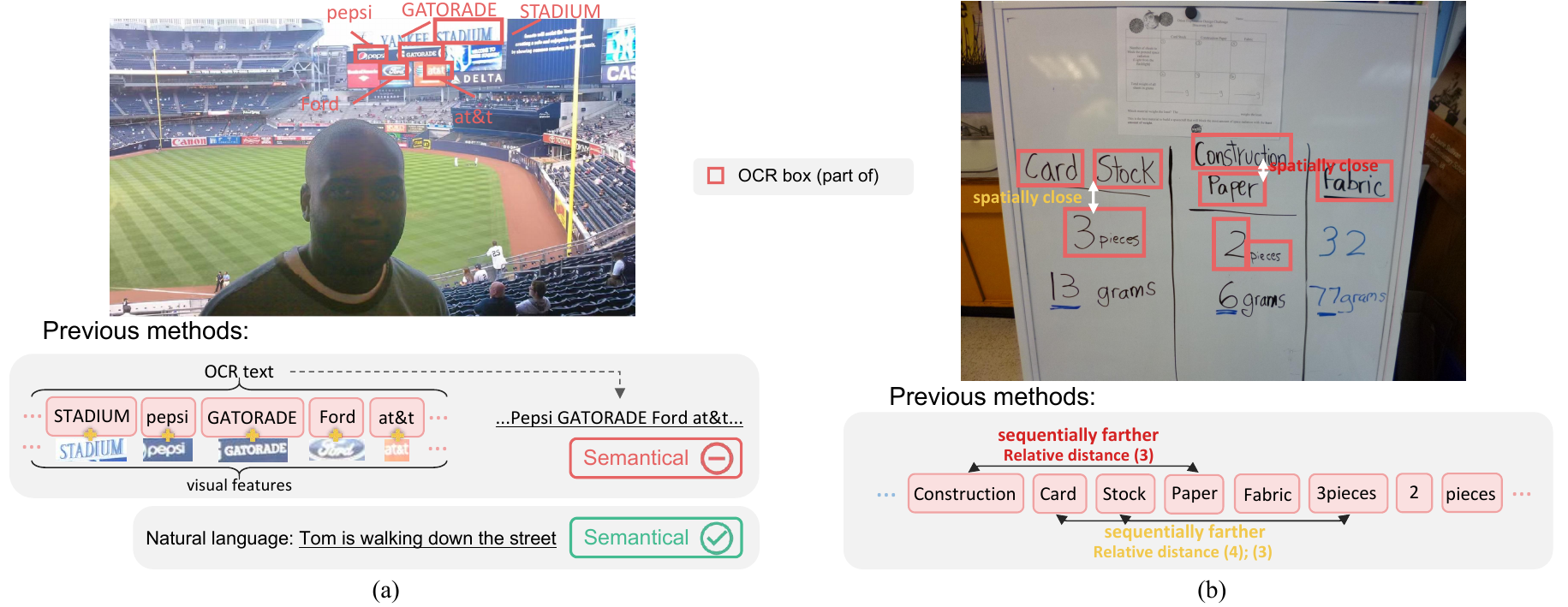}
  \caption{(a) Previous methods spliced the OCR text into a sentence according to the reading order. This sentence is not consistent with the sentence with the complete contextual semantic relationship input in the NLP task. There is no semantic relationship between many OCR texts, which brings noises. (b) Previous methods use 1-D position encoding according to the reading order, which cannot well represent spatial position relationships between OCR texts.}
  \label{figure_1}
  \Description{}

\end{figure*}

\section{Introduction}
Text-based Visual Question Answering (TextVQA) \cite{singh2019towards,biten2019scene,mishra2019ocr} aims to answer questions by understanding the interaction of scene text and visual information in the image. It requires models not only to recognize the difference between text and visual information but also to have the ability to reason the answer based on various modalities. Most of the works in this field explore the network structure \cite{kant2020spatially,hu2020iterative,zeng2021beyond,jin2022token,han2020finding,fang2022towards} and design specific pre-training tasks \cite{yang2021tap,biten2022latr,lu2021localize}, which promote field development. All of them use a transformer-based model \cite{vaswani2017attention}, which is trained from scratch or initiated by a  pre-training language model, to fuse different modality information. These transformer-based models \cite{devlin2018bert,raffel2020exploring} are usually used to process semantically coherent and complete sentences in Natural Language Processing (NLP).

However, the text in images is recognized by the Optical Character Recognition (OCR) system \cite{borisyuk2018rosetta}. These texts are scattered throughout the image. Compared with the input sentence in NLP tasks, \textbf{these OCR texts distributed in the image may not be semantically related and cannot form a sentence}. Figure \ref{figure_1} (a) shows the difference between the input of the TextVQA models and a natural language sentence. In Figure \ref{figure_1} (a), the sentence ``\texttt{Tom is walking down the street}'' is semantically coherent and complete. There are apparent contextual relationships in natural language sentences, and the words in the sentence have semantic relevance. Conversely, the OCR texts ``Pepsi GATORADE Ford at\&t'' do not actually have contextual semantic associations. However, all the previous works in TextVQA ignore that texts in an image are different from the sentence in NLP. The OCR texts sequence ``Pepsi GATORADE Ford at\&t'' is regarded as a sentence in current methods, even though the words in it do not actually have contextual semantic associations. Forcing these irrelevant OCR texts to form a sentence will force the model to construct contextual relationships that should not exist in these texts, adding harmful noise to the learning process of the model.

Another difference is that natural language text input has a reading order from left to right and top to down. The words and sentences in natural language inherently have semantic associations, so the text can be spliced in reading order and form as a linguistic sequence. There is no problem to use absolute position embedding or relative position embedding to indicate the 
sequential-position relationships between different words or sentences. 
However, OCR texts recognized in the scene image cannot simply be spliced from left to right and top to bottom. We attribute this to the nature that \textbf{OCR texts show strong spatial-position relations between each other}. The 1-D relative or absolute position encoding cannot express the 2-D complex spatial relationship in the image. It is not reasonable if we input the concatenated OCR texts into the model and then use the original 1-D position embedding for position modeling. This will lead to some OCR texts that are spatially close in the image being set away from each other by the 1-D position embedding. Intuitively, OCR texts that are adjacent left and right, or up and down in an image are more likely to have direct semantic associations. For example, in Figure \ref{figure_1} (b), current methods take in ``\texttt{Construction} \texttt{Card} \texttt{Stock} \texttt{Paper} \texttt{Fabric} \texttt{3pieces}...'' with 1-D position embedding adding to them, which will cause `\texttt{Construction}' to be close to `\texttt{Card}' while the distance between `\texttt{Construction}' and `\texttt{Paper}' is far. It will also lead to `\texttt{3 pieces}' close to `\texttt{Fabric}' but farther away from `\texttt{Card Stock}'. However, `\texttt{3 pieces}’ is more semantically related to `\texttt{Card Stock}' which is spatially close. Therefore, better spatial-aware position modeling is appealing. 

To alleviate the above problems, we rethink the text in Text-based Visual Question Answering and propose a new method called Separate and Locate (SaL).  Aiming at the problem that there is no apparent semantic association between the text in the image, we introduce a Text Semantic Separate (TSS) module. Compared with directly combining unrelated OCR texts into a sentence, TSS can learn to reduce the noise during training by separating OCR texts that do not have semantic relevance. In this way, the model is promoted to better learn relationships between different OCR texts and help the subsequent reasoning of answers to text-related questions. As for the problem that 1-D position encoding cannot properly express the spatial position relationships between OCR texts, we introduce a Spatial Circle Position (SCP) module. SCP provides each text in the image with representations indicating the spatial relative distances between it and all other texts. Specifically, we follow the Pythagorean theorem \cite{maor2019pythagorean} to calculate spatial relative distances between two texts.

\begin{figure*}[!htb]
  \centering
  \includegraphics[width=1\linewidth]{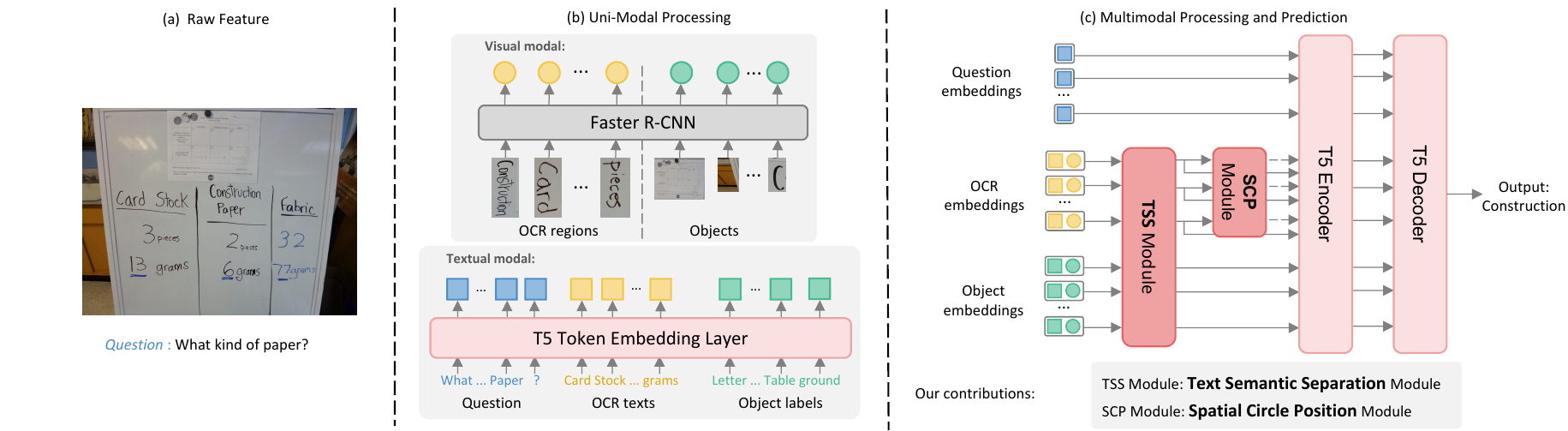}
  \caption{ The pipeline of our model. Same-shape markers represent features of the same modality, and marks of the same color represent features from the same text mark or image region.}
  \label{figure_2}
  \Description{}

\end{figure*}

Benefiting from the two modules, our proposed SaL properly overcomes the problems of ambiguous semantic associations of OCR texts and inaccurate positional representations. With better correlation capturing and the spatial-position realization between different OCR texts, SaL enhances the model's feature fusion ability and multi-modal information reasoning ability.

We conduct experiments on two benchmarks: TextVQA \cite{singh2019towards} and ST-VQA \cite{biten2019scene}. SaL, without any pretraining tasks that are adapted in previous works\cite{biten2022latr,lu2021localize,yang2021tap}, outperforms the SOTA methods even including those pre-training ones. The reason for not adapting pre-training is that OCR annotations of the SOTA pre-training dataset with 64 million samples are not open-source. We believe that our performance can be further improved by pre-training on the large-scale pre-training data.

In summary, our contributions are three-folded:

1. We are the first to claim that the text input in TextVQA is different from that in NLP. Most of the OCR texts do not have semantic associations and should not be stitched together. For this, we designed a Text Semantic Separate (TSS) module.

2. We propose a Spatial Circle Position (SCP) module to help the model realize the spatial relative-position information between OCR texts. It can solve the problem that 1-D position embedding cannot well represent the text position in the image.

3. Extensive experiments demonstrate the effectiveness of our method. SaL outperforms the current SOTA method by 2.68\% and 2.52\% on TextVQA and ST-VQA respectively (The SOTA method uses 64 million pre-training data but we do not use any pre-training data).

\section{Related Work}
\subsection{Vision-language tasks incorporating scene text}
As scene-text visual question answering \cite{biten2019scene,singh2019towards} has gradually gained attention, in order to enhance the scene-text understanding ability of VQA models \cite{antol2015vqa,alberti2019fusion}, several datasets \cite{singh2019towards,biten2019scene,mishra2019ocr} have been proposed to promote the development of this field.

Previous works \cite{liu2020cascade,singh2021textocr,wang2020general,han2020finding,hu2020iterative,fang2022towards,biten2022latr,wang2022tag,lu2021localize,yang2021tap,zeng2021beyond,jin2022token} realize that texts play an important role in answering text-related questions. CRN \cite{liu2020cascade} focuses on the interaction between text and visual objects. LaAP-Net \cite{han2020finding}  gives the bounding box of the generated answer to guide the process of answer generation. SMA \cite{gao2020structured} and SA-M4C \cite{kant2020spatially} build graphs to build relations between OCR texts and objects. TAG \cite{wang2022tag} introduces a data augmentation method for TextVQA.

TAP \cite{yang2021tap}, LOGOS \cite{lu2021localize}, LaTr \cite{biten2022latr}, and PreSTU \cite{kil2022prestu} propose different pre-training tasks to promote the development of scene-text visual question answering. Specifically, TAP proposed a text-aware pre-training task that predicts the relative spatial relation between OCR texts. LOGOS  introduces a question-visual grounding task to enhance the connection between text and image regions. LaTr is based on the T5 model \cite{raffel2020exploring} and proposes a layout-aware pre-training task that incorporation of the layout information. PreSTU is based on the mT5 model and designed a simple pre-training recipe for scene-text understanding. 

However, all of them ignore the irrelevance between different OCR texts and the poor ability of original 1-D position encoding. Our model separates  different OCR words according to their semantic contextual information and can realize the difference between OCR texts and complete semantically related sentences. Furthermore, it can establish accurate relative spatial relationships between each OCR word, which facilitates the answering process.
\subsection{Spatial position encoding methods}
After the Transformer \cite{devlin2018bert} becomes the common paradigm of NLP, the 1-D absolute position encoding \cite{devlin2018bert} and relative position encoding \cite{shaw2018self} are proposed to identify the position relation between the different words in the sentence. 
After Vit \cite{dosovitskiy2020image} uses the transformer to process the image task, Standalone self-attention \cite{ramachandran2019studying} proposes an encoding method for 2-D images. The idea is simple. It divides the 2-D relative encoding into horizontal and vertical directions, such that each direction can be modeled by a 1-D encoding.

However,  it only gives each image region a simple absolute spatial position encoding, and cannot directly construct the relative spatial relationship and distance between images. Simply summing two one-dimensional positional embeddings of x and y to represent the positional relationship of regions in an image is the main limitation that prevents the model from learning spatial relationships.

LaAP-Net \cite{han2020finding}, SMA \cite{gao2020structured}, SA-M4C \cite{kant2020spatially}, and LaTr \cite{biten2022latr} prove the critical role of the position of OCR texts in the TextVQA field. They proposed various network structures and pre-training tasks to make the model learn the different spatial position relations between OCR texts. LaAP-Net, SMA, and SA-M4C still use traditional 1-D absolute position encoding in NLP to build spatial associations between OCR texts.  Although LaTr uses six learnable lookup tables to construct the spatial layout information of OCR text, it does not improve much in the absence of a large amount of data pre-training, because it still uses  the simple addition of multiple original 1-D  position encoding to represent spatial position information.

Our method models the spatial relative distance and implicit angle of each OCR text in the image to all other OCR texts, which is a more direct and reasonable spatial encoding representation.

\section{Method}
SaL analyzes the difference between OCR texts in TextVQA and complete sentences in NLP. In terms of semantic relations, SaL proposes a Text Semantic Separate (TSS) module that explicitly separates OCR texts without semantic contextual relations.  In terms of spatial-position relations,  SaL introduces a Spatial Circle Position (SCP) module that models 2-D relative spatial-position relations for OCR texts. With these two modules, our method can separate semantic irrelevance OCR texts and locate the accurate spatial position of OCR texts in the image.

In this section, we introduce the whole process of our model. Specifically, we elaborate on the TSS and SCP modules in Sections 3.2 and 3.3 respectively.

\subsection{Multimodal Feature Embedding}
Following LaTr, we utilize a transformer of Text-to-Text Transfer Transformer (T5) as the backbone. As shown in Figure \ref{figure_2}(a), the original data in each sample in the dataset is a image and the corresponding question. Next, as shown in Figure \ref{figure_2}(b), we use the FRCNN model and the T5 token embedding layer to process visual and text information respectively. Finally, as shown in Figure \ref{figure_2}(c), the question, OCR text, and object features are concatenated together and input into the model. The specific process is as follows:

\textbf{Question Features.} Following LaTr, the question words are indexed as the question feature embeddings
\begin{math}
  Q=\left \{{q_{i}}\right \}_{i=1}^{L}
\end{math} by T5 token embedding layer, where  \begin{math}
  q_{i}\in \mathbb{R}^{d}
\end{math} 
is the embedding of the $i$-th question word, $L$ is the length of the question, and $d$ is the dimension of the feature.

\textbf{OCR Features.} For texts recognized in input images by the OCR system, we have three different features: 1) visual features extracted by Faster R-CNN $x_{i}^{ocr,fr}$. 2) the corresponding bounding box of visual features $x_{i}^{ocr,bx} = [\frac{x_{i}^{min}}{W},\frac{y_{i}^{min}}{H},\frac{x_{i}^{max}}{W},\frac{y_{i}^{max}}{H}]$. 3) the text embedding $x_{i}^{ocr,t5}$ produced by the T5 embedding layer. The final OCR feature is: 
\begin{equation}
      x_{i}^{ocr}=T5LN (W_{fr}x_{i}^{ocr,fr}) +T5LN (W_{bx}x_{i}^{ocr,bx}) + W_{t5}x_{i}^{ocr,t5}  \\
\end{equation}

\textbf{Object Features.} To get object region features, we apply the same Faster R-CNN model mentioned in OCR features:
\begin{equation}
      x_{i}^{obj}=T5LN (W_{fr}^{'}x_{i}^{obj,fr}) +T5 LN (W_{bx}^{'}x_{i}^{obj,bx}) + W_{t5}^{'}x_{i}^{obj,t5}  \\
\end{equation} where \begin{math}x_{i}^{obj,fr}\end{math} is the appearance feature, \begin{math}x_{i}^{obj,bx}\end{math} is the bounding box feature and \begin{math}x_{i}^{obj,t5}\end{math} is the t5 word embedding corresponding to the object label.

Therefore, the input embedding is:
\begin{equation}
      input = cat (X^{q},X^{ocr},X^{obj})  \\
\end{equation}
where $X^{q}$ is the question embeddings, $X^{ocr}$ is the OCR embeddings, $X^{obj}$ is the object embeddings. The \textit{cat} means the concatenating function. 

\begin{figure*}[!htb]
  \centering
  \includegraphics[width=1\linewidth]{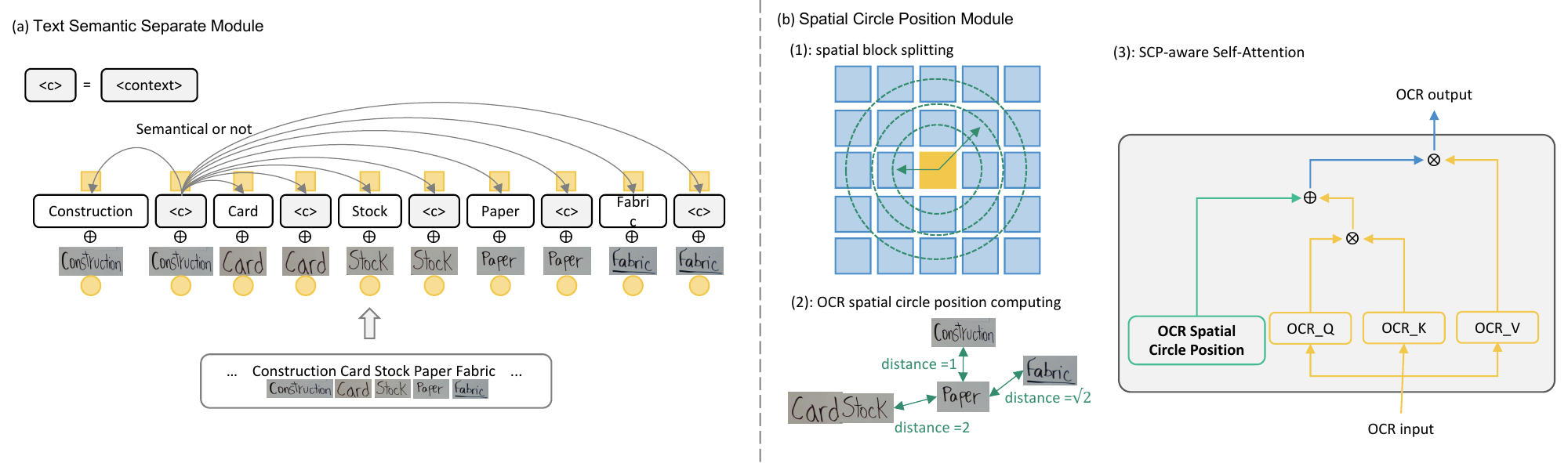}
  \caption{\textbf{(a) Text Semantic Separation module}: OCR texts are split by the <context> token which will learn whether two adjacent tokens are semantic-related. \textbf{(b) Spatial Circle Position module} (b1) splits the image into blocks, (b2) computes the spatial circle distances between blocks at which the OCR texts are located, (b3) and then adds the corresponding position embedding into the SCP-aware Self-Attetion. }
  \label{figure_3}
  \Description{}
\end{figure*}

\subsection{Text Semantic Separate Module}
Unlike words in a sentence, scene texts in images are distributed in various positions of the image. They are distributed on different text carriers and backgrounds of different materials, which naturally leads to no contextual relationship between most texts. Since the previous work did not realize this, OCR texts are directly spliced as a sentence and input into the model. This makes the learning process of the model suffer from noises. 

To this end, the Text Semantic Separate (TSS) module separates different OCR texts according to their visual and spatial information, so that the model can correctly recognize differences between OCR texts and the natural sentence, and better help the model to express and fuse the features of the question, the OCR text, and the object. Specifically, we can look at the bottom of Figure \ref{figure_3}(a). Without the TSS module, the model stitches the OCR text together, making the model think that two adjacent words such as `\texttt{Paper}' and `\texttt{Fabric}' may have contextual relationships. However, it is not the case. 

Our TSS module inserts a context-separation token 
<context> after the last token of each OCR text. Every <context> token is represented by its learnable reserved embedding from the T5 lookup table. Then, we add the visual feature and bounding coordination of each OCR text to its corresponding context-separation token (shown in Figure \ref{figure_3} (a)). Finally, each context-separation token can interact with all other OCR texts and distinguish whether there is a semantic relationship by visual relation and coordination relation. 
The benefits of this are: 1) The model can learn that there is no contextual relationship between different OCR texts, which helps the model's reasoning and feature fusion. 2) <context> can combine the OCR texts before and after to learn the difference between different OCR texts. 3) Compared with directly splicing OCR text into sentences, this method reduces noise.

\subsection{Spatial Circle Position Module}
The spatial position relationship of text in natural scenes is extremely important for solving text-related questions. How to construct the complex spatial position relationship between texts has become one of the urgent problems to be solved. We, therefore, propose the SCP module to construct precise inter-textual spatial position 
relationships. 

Specifically, SCP includes the following three steps: 1) divide the scene image into 11 $*$ 11 image areas, and assign all OCR texts to the corresponding areas by their coordinates; 2) calculate the spatial distance between each OCR text and all other texts through the Pythagorean theorem; 3) assign spatial position embeddings between the OCR text and other OCR texts based on the calculated spatial distances and feed them into the spatial circle position-aware self-attention. The formula for this process is as follows:

\begin{flalign}
& \qquad \qquad p_{i}^{ocr} = Patch(x_{i}^{ocr,bx})  \\
& \qquad \qquad dist_{i,j}^{ocr} = (Pytha(p_{i}^{ocr},p_{j}^{ocr})*2).long()\\
& \qquad \qquad distEmbed_{i,j}^{ocr}= Embedding(dist_{i,j}^{ocr})\\
& \qquad \qquad att_{i,j} = \frac{ (W_{q}*ocrInput)  (W_{k}*ocrInput) ^{T} + b_{i,j}}{\sqrt{d}}\\
& \qquad \qquad \alpha_{i,j}=softmax(att_{i,j}+distEmbed_{i,j}^{ocr}) \\
& \qquad \qquad ocrOutput_{i,j} = \alpha_{i,j}*(W_{v}*ocrInput)&
\end{flalign}

where Patch is a function that aligns the OCR text to the image patch coordination by its coordination. Pytha is a function that calculates the spatial distance between OCR texts.
Embedding is a function in Pytorch and it is a 32*12 look-up table. \begin{math}
  W_{q}, W_{k}, W_{v}
\end{math} are learnable parameters of the self-attention layer and \begin{math}
  \frac{1}{\sqrt{d}} 
\end{math} is a scaling factor.

Using the SCP module has the following advantages: 1) Compared with models such as SMA and SA-M4C, which construct an attention layer that specifically handles a variety of predefined spatial position relationships. SCP explicitly constructs all spatial relationships of each OCR text with other OCR texts through spatial distance. In addition, since in the first step of Figure \ref{figure_3} (b), various angle relationships implicitly exist between each image block, and each OCR text assigned to a different image block also implicitly contains various angle relationships. It does not require additional consumption of model parameters, only a 32*12 learnable look-up table is required.
2) Compared with LaTr, which uses 6 learnable look-up tables of 1000*768 and a large amount of pre-training data, the explicit construction method of SCP is much better (See Table \ref{table5} for details).
3) The previous method builds global spatial positional relationships of all OCR texts and cannot well construct the relative spatial positional relationship between each OCR text and all other OCR texts. SCP takes into account the implicit angle and spatial distance between each OCR text and all other OCR texts, and can more accurately locate the position of the text in the image.

As we can see in Figure \ref{figure_3} (a), the previous methods tend to generate many unreasonable positional relationships due to the lack of spatial position representation capabilities. For example, `\texttt{Construction}' is closer to `\texttt{Card}' but `\texttt{Construction}' is farther away from `\texttt{Paper}'. This obviously doesn't match what we see in the image. Our approach solves this problem very well. Specifically, as shown in Figure \ref{figure_3} (b3), we added our SCP module to the attention layer, so that the model can capture various angles and spatial position relationships between OCR texts. 

\subsection{Training Loss}
Following previous works, we use the binary cross-entropy loss to train our model. Since there are several answers to the question in the sample, it can be converted into a multi-label classification problem. The binary cross-entropy loss can be defined as followings:
\begin{equation}
    \begin{split}
        pred &= \frac{1}{1+exp (-y_{pred}) }\\
        L_{bce} &= -(y_{gt}log (pred)  +  (1-y_{gt}) log (1-pred)) 
    \end{split}
\end{equation}
where $y_{pred}$ is the prediction and $y_{gt}$ is the ground-truth target.

\begin{table*}[t]
  \caption{Comparison on the TextVQA dataset. For a fair comparison, the top of the table is the result of using Rosetta OCR and only using the TextVQA dataset training. The bottom of the table is the result of the unrestricted setting. $\ddagger$ to refer to the models trained on TextVQA and ST-VQA.
  }

  \label{table1}

  \begin{tabular}{ccccccc}
    \toprule
    \# &Model &OCR system&Pre-Training Data&Extra Finetune & Val Acc. (\%)   & Test Acc. (\%)   \\
  
    \midrule
    1&M4C \cite{hu2020iterative} & Rosetta-en & -  &-&  39.40&  39.01\\
    2&SMA \cite{gao2020structured} & Rosetta-en& - &-&  40.05 &  40.66\\
    3& LaAP-Net \cite{han2020finding} & Rosetta-en&  - &-& 40.68 &  40.54\\
    4&CRN \cite{liu2020cascade} & Rosetta-en& - &-&  40.39 &  40.96\\
    5&BOV \cite{zeng2021beyond} & Rosetta-en& - &-&  40.90 &  41.23\\
    6&SC-Net \cite{fang2022towards} & Rosetta-en& - &-&  41.17 &  41.42\\
    7&TAP \cite{yang2021tap} & Rosetta-en& TextVQA &-&  44.06 &  -\\
    8&LaTr-Base \cite{biten2022latr} & Rosetta-en& - &-&  44.06 &  -\\
    \rowcolor{gray!30}
    9&SaL-Base & Rosetta-en& - &-&  \textbf{48.67} &  -\\
   
    \hline
    
    10&SA-M4C \cite{kant2020spatially} & Google-OCR& - &ST-VQA&  45.40 &  44.60\\
    11&SMA \cite{gao2020structured} & SBD-Trans OCR& - &ST-VQA&  - &  45.51\\
    12&LOGOS \cite{lu2021localize} & Microsoft-OCR& - &ST-VQA&  51.53 &  51.08\\
    13&TWA \cite{jin2022token} & Microsoft-OCR& - &ST-VQA&  52.70 &  52.40\\
    14&TAG \cite{wang2022tag} & Microsoft-OCR& TextVQA,ST-VQA &ST-VQA&  53.63 & 53.69\\
    15&TAP \cite{yang2021tap} & Microsoft-OCR& TextVQA,ST-VQA,TextCaps,OCR-CC &ST-VQA&  54.71 &  53.97\\
    16&LaTr-Base \cite{biten2022latr} & Amazon-OCR& IDL &-& 58.03 &  58.86\\
    17&LaTr$\ddagger$-Base \cite{biten2022latr} & Amazon-OCR& IDL &ST-VQA& 59.53 &  59.55\\
    18&LaTr$\ddagger$-Large \cite{biten2022latr} & Amazon-OCR& IDL &ST-VQA& 61.05 &  61.60\\
    \rowcolor{gray!30}
    19&SaL-Base & Amazon-OCR& -  &-&  62.42&  62.35\\
    \rowcolor{gray!30}
    20&SaL$\ddagger$-Base & Amazon-OCR& -  &ST-VQA&  62.85&  63.18 \\
    \rowcolor{gray!30}
    21&SaL-Large   & Amazon-OCR&  -  & -& 63.88&  63.92\\
    \rowcolor{gray!30}
    22&SaL$\ddagger$-Large   & Amazon-OCR& - & ST-VQA& \textbf{64.58}& \textbf{64.28}\\
  
    \bottomrule
  \end{tabular}
\end{table*}

\section{Experiments}
In this section, we verify the effectiveness of our method on two main benchmarks, TextVQA and ST-VQA. In Section 4.1, we introduce them and evaluation metrics. 

In Sections 4.2 and 4.3, we compare our method with SOTA methods and conduct ablation experiments. Finally, in Section 4.4 we present the visualization results. Following LaTr, we use $\ddagger$ to refer to the models trained on TextVQA and ST-VQA. `-Base' and `-Large' model sizes refer to architectures that have 12+12 and 24+24 layers of transformers in encoder and decoder, respectively. 

\subsection{Datasets and Evaluation Metrics}
\textbf{TextVQA} \cite{singh2019towards} is the main dataset for text-based visual question answering.  It is a subset of open images \cite{kuznetsova2020open} and is annotated with 10 ground-truth answers for each scene-text-related question. It includes 28,408 images with 45,336 questions. Following previous settings, we split the dataset into 21,953 images, 3,166 images, and 3,289 images respectively for train, validation, and test set. The methods are evaluated by the soft-voting accuracy of 10 answers.

\textbf{ST-VQA} is similar to TextVQA and it contains 23,038 images with 31,791 questions. We follow the setting from M4C and split the dataset into train, validation, and test splits with 17,028, 1,893, and 2,971 images respectively. The data in ST-VQA is collected from Cocotext \cite{COCO-Text}, Visual Genome \cite{Genome}, VizWiz \cite{VizWiz}, ICDAR \cite{karatzas2015icdar2015,icdar2013}, ImageNet \cite{ImageNet}, and IIIT-STR \cite{MishraAJ13} datasets. Different from TextVQA, we report both soft-voting accuracy and Average Normalized Levenshtein Similarity(ANLS) on this dataset. ANLS  defined as \begin{math}1-d_{L}  (pred,gt) /max (|pred|,|gt|)   \end{math}, where \begin{math}pred \end{math} is the prediction, \begin{math} gt \end{math}  is the ground-truth answer, \begin{math} d_{L} \end{math} is the edit distance.

\subsection{Experiment Results}

\subsubsection{TextVQA Results}
For a fair comparison with previous methods, we divide the previous methods into non-pre-trained methods and pre-trained methods pre-training on numerous data. Since the accuracy of the OCR system for recognizing text has a great influence on the performance of the model, our method uses the classic Rosetta OCR and Amazon OCR to make a fairer comparison with the previous methods.

As shown in Table \ref{table1}, in the case of using Rosetta OCR, the performance of our method has reached an accuracy rate of 48.67\% in the validation set of TextVQA, which exceeds the previous SOTA method LaTr by 4.61\%.

In the case of using Amazon-OCR, our method uses the same T5-base as the previous SOTA LaTr-base as the model structure, and our method outperforms LaTr by 3.32\% and 3.63\% accuracy on the validation set and test set of TextVQA, respectively. In the case of using the T5-large and training on both TextVQA and ST-VQA datasets, our method achieved the new best accuracy of 64.58\% and 64.28\% on the validation set and test set of TextVQA respectively. Our method outperforms the SOTA LaTr-large method with the same configuration by 3.53\% and 2.68\% respectively. It is worth noting that the  additional pre-training data set IDL \cite{biten2022latr} containing 64 million data used by LaTr is not open source, which hinders our method from pre-training. This demonstrates the effectiveness and efficiency of our method.  Since this pre-training method is orthogonal to our method, we believe that our method will be further improved by applying the pre-training task of LaTr.

\subsubsection{ST-VQA Results}
As we can see in Table \ref{table2}, under the unconstrained setting, SaL$\ddagger$-Large achieves 64.16\% accuracy, 0.722 ANLS on the validation set of ST-VQA, and ANLS of 0.717 on the test set. Our model exceeds SOTA LaTr$\ddagger$-Large by 2.52\% accuracy, 2.0\% ANLS on the ST-VQA validation set, and 2.1\% ANLS on the ST-VQA test set. Likewise, SaL-base also outperforms LaTr-base by a large margin. From Table \ref{table1} and Table \ref{table2}, we can observe that training on both TextVQA and ST-VQA, the improvement on TextVQA (from 62.42\% to 62.85\% on SaL-base, from 63.88\% to 64.58\% on SaL-large ) is not significant as the improvement on the ST-VQA dataset (from 59.74\% to 62.29\% on SaL-base and from 61.45\% to 64.16\% on SaL-large). We will  analyze this phenomenon in  Appendix B.


\begin{table}[t]
  \caption{Results on the ST-VQA Dataset. SaL outperforms the state-of-the-art by 2.52\% accuracy without extra pretraining data. }

  \label{table2}

  \begin{tabular}{ccccccc}
    \toprule
    \# &Model & Val Acc. (\%)  & Val ANLS  & Test ANLS  \\
  
    \midrule
    1&M4C \cite{hu2020iterative} &   38.05& 0.472& 0.462 \\
    2&SA-M4C \cite{kant2020spatially} &   42.23& 0.512& 0.504 \\
    3&SMA \cite{gao2020structured} &  -&-& 0.466 \\
    4&CRN \cite{liu2020cascade} &  -&-& 0.483 \\
    5&LaAP-Net \cite{han2020finding} &  39.74&0.497& 0.485 \\
    6&LOGOS \cite{lu2021localize} &  48.63&0.581& 0.579 \\
    7&TAP  \cite{yang2021tap}&  50.83&0.598& 0.597 \\
    8&LaTr-Base \cite{biten2022latr} &  58.41&0.675& 0.668 \\
    9&LaTr$\ddagger$-Base \cite{biten2022latr} &  59.09&0.683& 0.684 \\
    10&LaTr$\ddagger$-Large \cite{biten2022latr} &  61.64&0.702& 0.696 \\

    \rowcolor{gray!30}
    11&SaL-Base &  59.74&  0.683&  0.673\\
    \rowcolor{gray!30}
    12&SaL$\ddagger$-Base & 62.29 &0.708&  0.697 \\
    \rowcolor{gray!30}
    13&SaL-Large  &61.45& 0.699&  0.685\\
    \rowcolor{gray!30}
    14&SaL$\ddagger$-Large  & \textbf{64.16}& \textbf{0.722}&\textbf{ 0.717}\\
  
    \bottomrule
  \end{tabular}
\end{table}

\subsection{Ablation Study}
In this section, we provide insightful experiments that we deem influential for the TextVQA task and its future development. 
We first analyze the performance improvement of the different modules we propose on the TextVQA dataset. Afterward, we analyze the effect of different input information on the model performance. Then, to prove the effectiveness of our SCP module, we compare our SCP module and layout embedding in LaTr. Finally, for different methods of OCR text separation, we conduct experiments and further emphasize that our TSS module is the most effective choice.

\begin{table}[]
\caption{Ablation studies of different modules on TextVQA and ST-VQA datasets. TSS and SCP present text semantic separate module and spatial circle position  module, respectively. }
\label{table3}
\begin{tabular}{llllll}

\toprule
\multirow{2}{*}{\#} & \multirow{2}{*}{Model} & \multirow{2}{*}{Module} & \multirow{2}{*}{\begin{tabular}[c]{@{}l@{}}TextVQA\\ Acc.(\%)\end{tabular}} & \multirow{2}{*}{\begin{tabular}[c]{@{}l@{}}ST-VQA\\ Acc.(\%)\end{tabular}} & \multirow{2}{*}{\begin{tabular}[c]{@{}l@{}}ST-VQA\\ ANLS\end{tabular}} \\ 
                    &  &  &  &   &  \\ 
\midrule
1  & Baseline & - & 57.98  & 55.78 & 0.649 \\
2  & Baseline & TSS & 61.55 & 58.29 & 0.667 \\
3  & Baseline & SCP  & 60.98 & 57.80 & 0.666  \\
4  & Baseline & TSS + SCP  & 62.42 & 59.74  & 0.683 \\  
\bottomrule
\end{tabular}
\end{table}

\subsubsection{The Effective Of Modules}
In order to prove the effectiveness of our proposed method, we follow LaTr to do ablation experiments on TextVQA and ST-VQA based on T5-Base. 

As shown in the first row of Table \ref{table3}, our baseline model (removing all of our proposed modules) shows the worst performance compared to our full model and other ablated ones. 

As shown in the second row of Table \ref{table3}, with the help of the TSS module, the accuracy of our baseline on TextVQA increased from 57.98\% to 61.55\%, and the accuracy on ST-VQA increased from 55.78\% to 58.29\%. This proves that the TSS module can make the model realize whether there is a semantic context relationship directly between different OCR texts, reducing the noise generated by treating all OCR texts as a sentence in previous work. 

When adding the SCP module, the performance of the baseline on TextVQA and ST-VQA increased from 57.98\% to 60.98\% and from 55.78\% to 57.80\% respectively. At the same time, the ANLS metric also improved by 1.7\% on ST-VQA. This proves that the SCP module can help the model better locate OCR texts in different spatial positions in the image, and construct the spatial relationship between OCR texts in different positions. 

Adapting TSS and SCP modules at the same time (our full model), compared with the baseline, the accuracy on TextVQA and ST-VQA is increased by 4.44\% and 3.96\% respectively. The ANLS of our model on ST-VQA maintained the same trend as the accuracy and increased by 3.40\%, indicating the importance of both modules. In the visualization section, we will more intuitively show the impact of these two modules on model inference.

\begin{table}[]
\caption{Ablation studies of different input information.}
\label{table4}
\scalebox{0.9}{
\begin{tabular}{cllllll}
\toprule
\multicolumn{1}{l}{\multirow{2}{*}{Model}} & \multirow{2}{*}{Quetsion} & \multirow{2}{*}{\begin{tabular}[c]{@{}l@{}}OCR \\ Text\end{tabular}} & \multirow{2}{*}{\begin{tabular}[c]{@{}l@{}}OCR \\ Visual\end{tabular}} & \multirow{2}{*}{\begin{tabular}[c]{@{}l@{}}OBJ\\ Label\end{tabular}} & \multirow{2}{*}{\begin{tabular}[c]{@{}l@{}}OBJ\\ Visual\end{tabular}} & \multirow{2}{*}{Acc.} \\
\multicolumn{1}{l}{}                       &                           &                                                                      &                                                                        &                                                                      &                                                                       &                       \\ \midrule
\multirow{6}{*}{SaL} &  \CheckmarkBold  &  \XSolidBrush &  \XSolidBrush &  \XSolidBrush &  \XSolidBrush  & 11.97 \\
   &  \CheckmarkBold &  \CheckmarkBold &  \XSolidBrush   &  \XSolidBrush   &  \XSolidBrush & 56.17   \\
  &  \CheckmarkBold &  \CheckmarkBold & ViT   &  \XSolidBrush   &  \XSolidBrush  & 56.78 \\
 &  \CheckmarkBold  &  \CheckmarkBold   & FRCNN &  \XSolidBrush  &  \XSolidBrush & 61.31  \\
&  \CheckmarkBold  &  \CheckmarkBold & FRCNN   &  \CheckmarkBold  &  \XSolidBrush  & 61.97 \\
 &  \CheckmarkBold  &  \CheckmarkBold& FRCNN  &  \CheckmarkBold & \CheckmarkBold &62.42 \\\bottomrule
\end{tabular}}
\end{table}

\subsubsection{The Influence Of Different Input Information}
We explore the effect of different input information on the model. We use the coordinate information of OCR and objects in the image by default. The results are shown in Table \ref{table4}. 
It can be seen that OCR information (including both text and visual information) has the greatest impact on the performance of the model. Adding OCR text and its visual information extracted by FRCNN achieves an accuracy of 61.31\%, which drastically improve the model with only question input.
In addition, different from the conclusion in LaTr, when using the same T5-Base model as the model structure, the performance of OCR using FRCNN visual features is much better than using ViT visual features in our experiments. 
For this, two reasons can be regarded. The first one is that compared to ViT, we bind FRCNN features with OCR text features through addition operations. This means that for each OCR text, we can accurately fuse its text and visual features. However, using ViT visual features requires the model to be trained to match each OCR text with its associated image patch. There is no supervisory signal to help OCR text to match ViT features during training, making the performance of using ViT visual features much worse than that of using FRCNN features. The second reason is that the ViT model needs to resize the image to 224*224, which greatly reduces the resolution of the image, making the ViT feature unable to express the visual information of the original image well.

\begin{figure*}[!htb]
  \centering
  \includegraphics[width=0.9\linewidth]{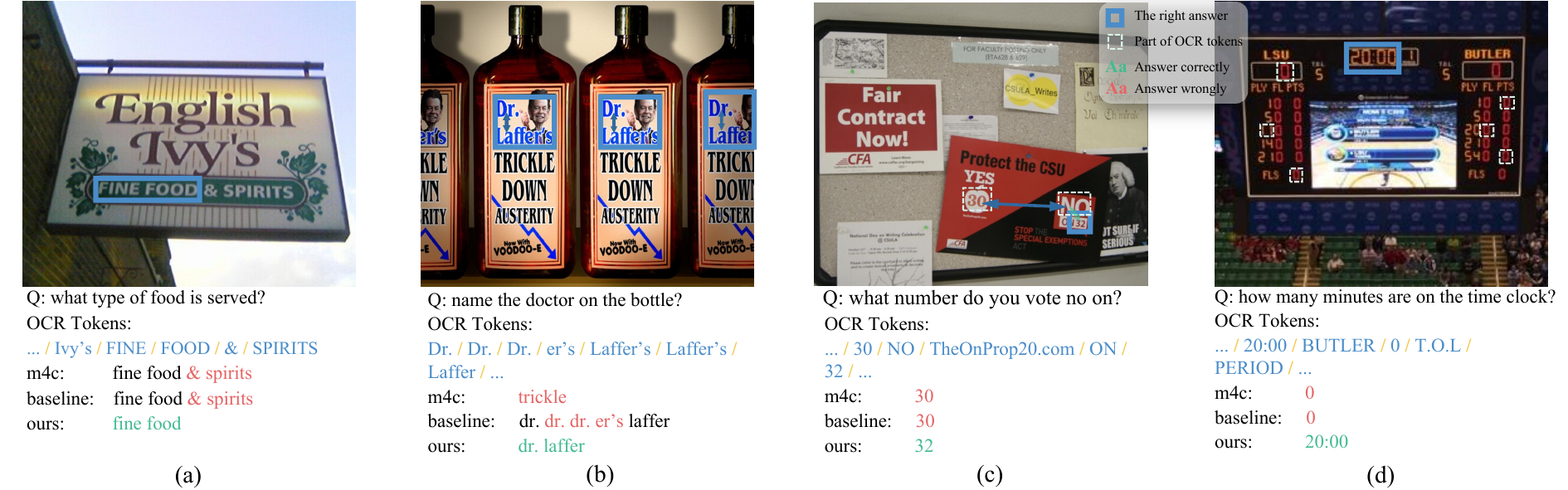}
  \caption{Some cases of SaL compared to the baseline and M4C. SaL can distinguish whether there is a contextual relationship between OCR texts and can better model the spatial position relationship between each OCR text and other OCR texts.}
  \label{figure_4}
  \Description{}

\end{figure*}

\begin{table}[h]
\caption{Effectiveness of  spatial circle position module. $\circ$ represent only use input question and OCR text.}
\label{table5}
\begin{tabular}{llllll}

\toprule
\# &Model &Module  &Val Acc. (\%)    \\
\midrule
1  & LaTr $\circ$ \cite{biten2022latr} & - & 50.37\%   \\
2  & LaTr $\circ$ \cite{biten2022latr} & layout embedding & 51.22\% \\
3  & SaL $\circ$ &  -  & 53.51\%  \\
4  & SaL $\circ$ &  layout embedding  & 54.56\%  \\
5  &  SaL $\circ$ & SCP  & 55.95\% \\  
\bottomrule
\end{tabular}
\end{table}

\subsubsection{Effectiveness Of  Spatial Circle Position Module. }
In order to further prove the importance of the relative spatial position and distance between the OCR texts in the image and the effectiveness of our SCP module, we compare the LaTr method that represents the absolute spatial position of the OCR text with our SCP module. For a fair comparison, both our model and LaTr only input questions and OCR text (no visual features). As shown in Table \ref{table5}, the use of layout embedding to represent the spatial position of the OCR text in the image increases the accuracy of the LaTr model in the TextVQA validation set by 0.85\%. We use layout embedding in SaL, and the accuracy increased from 53.51\% to 54.56\%. In contrast, when the SCP module is used to represent the relative spatial position relationship between each OCR text in the image, the accuracy of SaL is increased from 53.51\% to 55.95\%.

\begin{table}[h]
\caption{Ablation studies of different OCR text separation methods. }
\label{table6}
\begin{tabular}{llllll}

\toprule
\# &Model &Method  &Val Acc. (\%)    \\
\midrule
1  & Baseline & - & 57.98   \\
2  & Baseline & Tag & 59.74 \\
3  & Baseline &  Index & 60.07  \\
4  & Baseline & TSS  & 61.55 \\  
\bottomrule
\end{tabular}
\end{table}

\subsubsection{Different OCR Text Separation Methods}
This subsection studies the effect of different OCR text separation methods. We implement two method variants: Tag and Index. Tag add <context> and OCR visual feature to the last token of every OCR text in the image to  distinguish each OCR text  instead of  separating possible phrases in OCR texts. As for Index, it separates each OCR text by directly shifting its position id for 1-D position encoding instead of inserting <context>. 
The Tag variant provides the model with an embedding that learns the context between OCR texts, and the Index variant tells the model that the distance between different OCR texts should be appropriately distanced. As shown in Table \ref{table6}, both variants improve the performance, which further emphasizes our motivation of separating OCR texts. Compared with these variants, Our TSS achieves the best performance, indicating its superiority, as our TSS can satisfy the goals of both variants.

\subsection{Visualization}
To further verify the effectiveness of our method, we illustrate some visual cases from the TextVQA validation set. As shown in Figure \ref{figure_4} (a), since neither the m4c model nor the baseline model considers whether the OCR text has semantic relevance, OCR texts are directly processed as a sentence. Therefore, the m4c model and the baseline take the text ``\texttt{fine} \texttt{food} \& \texttt{spirits}'' belonging to the same line as the answer. With the help of the TSS module, our model learned whether there is a semantic relationship between each OCR text, and gave the correct answer `\texttt{fine food}'. It can be seen from Figure \ref{figure_4} (c, d) that for images with multiple OCR texts, our model can better model the spatial position relationship between them, and get the correct answer based on reasoning. Specifically, in Figure \ref{figure_4}(c), the spatial position relationship between `\texttt{32}' and `\texttt{NO}' is closer, and the baseline uses `\texttt{30}' as the answer as it is closer to `\texttt{NO}' according to the reading order. Figure \ref{figure_4} (b) can better reflect the role of our TSS module and SCP module. It can be seen that the baseline model directly stitches ``\texttt{dr.} \texttt{dr.} \texttt{dr.} \texttt{er's} \texttt{affer}'' together as the answer in accordance with the reading order, while our model takes the text into account semantics and the spatial positional relationship between OCR texts to give the correct answer. More  qualitative examples and error cases can find out  in  Appendix C.

\section{Conclusion}
For the TextVQA community, we proved that the previous works are unreasonable for OCR text processing. Previous works added noise to the reasoning process of the model by forcibly splicing all OCR texts into a sentence. 
Our proposed Text Semantic Separate module effectively separates different OCR texts and enables the model to learn whether there is a semantic context relationship between different OCR texts.
 In addition, the spatial position relationship of OCR text is very important for the model to understand the spatial relationship of OCR text in different positions in the image. But the 1-D position embedding used in previous work cannot reasonably express this relationship. The Spatial Circle Position module we proposed can better reflect the spatial position relationship between each OCR text and other OCR texts in the image. It can make the model more accurately locate the position of the OCR text in the image. 
  With these two modules, our proposed SaL model achieves SOTA performance on TextVQA and ST-VQA datasets without any pretraining tasks.
 Finally, we call on the community to rethink how to use the text information in the scene more reasonably.

\bibliographystyle{ACM-Reference-Format}
\balance
\bibliography{sample-base}

\appendix

\clearpage
\section{More Implementation Details}

In this section, we summarize the implementation and hyper-parameter settings of the  model designed in our work. All experiments are based on PyTorch deep-learning framework.

SaL consists of three main components: multimodal inputs,  a multi-layer transformer encoder, and a multi-layer transformer decoder. During training, the answer words are iteratively predicted and supervised using teacher-forcing. We apply multi-label sigmoid loss over the T5 tokenizer vocabulary score. We list the network and optimization parameters of SaL in Table \ref{table7} and Table \ref{table8} respectively.

\begin{table}[h]

  \caption{The network parameters of SaL. }
  \begin{tabular}{ll}
  
    \toprule
     Network parameters&Value\\\hline

    \midrule
   Maximum question tokens&{45}\\
    Maximum OCR tokens&{350}\\
    Maximum object tokens&{250}\\
    Text Faster R-CNN feature dimension&{2048}\\
    Object Faster R-CNN feature dimension&{2048}\\
    Joint embedding dimension&{768}\\
    Transformer encoder layers&{12}\\
    Transformer decoder layers&{12}\\
    Transformer attention heads&{12}\\
    Maximum decoding steps&{25}\\
    \bottomrule
  \end{tabular}

  \label{table7}
\end{table}

\begin{table}[h]

  \caption{The optimization parameters of SaL. }
  \begin{tabular}{ll}
  
    \toprule
     Optimization parameters&Value\\\hline

    \midrule
   Optimizer&{Adam}\\
    Base learning rate&{2e-4}\\
    Learning rate decay&{0.1}\\
    Warm-up iterations&{1000}\\
    Warm-up learning rate factor&{0.2}\\
    Batch size&{36}\\
    Max iterations&{24000}\\
    Learning rate decay steps&{14000, 19000}\\
    \bottomrule
  \end{tabular}

  \label{table8}
\end{table}

\begin{table}[]
\caption{Analysis of the model performances on TextVQA and ST-VQA. Len. represent the length of the answer. We count all answer lengths in each sample (10 answers).}
\label{table9}
\begin{tabular}{lcllll}
\toprule
\multicolumn{1}{c}{}                        &                                               & \multicolumn{3}{c}{Train Set}                                  & Val Set                       \\ \cline{3-6} 
\multicolumn{1}{c}{\multirow{-2}{*}{Model}} & \multirow{-2}{*}{Dataset}                     & Len.                 & Num     & Ratio                         & Acc.                          \\ \midrule
                                            & \multicolumn{1}{l}{}                          & $\le$ 3 & 324,963 & 0.939                         & 64.50                         \\
                                            & \multicolumn{1}{l}{\multirow{-2}{*}{TextVQA}} & $\textgreater $ 3  & 21,057  & 0.061                         & 51.71                         \\
                                            &                                               & $\le$ 3 & 549,373 & \cellcolor[HTML]{C0C0C0}0.946 & \cellcolor[HTML]{C0C0C0}65.26 \\
                                            & \multirow{-2}{*}{Both}                        & $\textgreater $ 3  & 31,107  & \cellcolor[HTML]{C0C0C0}0.054 & \cellcolor[HTML]{C0C0C0}49.17 \\ \cline{2-6} 
                                            &                                               & $\le$ 3 & 224,410 & 0.957                         & 62.27                         \\
                                            & \multirow{-2}{*}{ST-VQA}                      & $\textgreater $  3  & 10,050  & 0.043                         & 38.20                         \\
                                            &                                               & $\le$ 3 & 549,373 & \cellcolor[HTML]{C0C0C0}0.946 & \cellcolor[HTML]{C0C0C0}64.67 \\
\multirow{-8}{*}{SaL$\ddagger$-large}       & \multirow{-2}{*}{Both}                        & $\textgreater $ 3  & 31,107  & \cellcolor[HTML]{C0C0C0}0.054 & \cellcolor[HTML]{C0C0C0}46.67 \\ \bottomrule
\end{tabular}
\end{table}
\section{Why the performance is different on TextVQA and ST-VQA? }
 As for the observation of different improvement ranges on TextVQA and ST-VQA when SaL is jointly finetuned on them, we analyzed the accuracy of answers of different lengths in the two datasets to explain this phenomenon.
From Table \ref{table9}, we can see that in TextVQA and ST-VQA, the number of answers with lengths greater than 3 is far less than the number of answers with lengths less than or equal to 3. 
This makes the model tend to predict an answer with a length less than or equal to 3 and predict answers with smaller lengths more accurately. We can see that the ratio of answers with lengths of less than 3 to all answers in TextVQA is 0.939, and it reaches 0.957 in ST-VQA. This bias triggers the larger difference between the accuracy of different-length answers in ST-VQA than in TextVQA (62.27\% vs 38.20\% and 64.50\% vs 51.71\%).

When jointly finetuning on both TextVQA and ST-VQA, it can be observed from Table \ref{table9} that the proportion of answers with lengths greater than or equal to 3 becomes 0.946. This is a higher rate than in TextVQA but lower than that in ST-VQA. This means that the joint training makes the bias in TextVQA more severe but alleviates the bias in the ST-VQA.
Therefore, the accuracy of the answers of different lengths in the ST-VQA validation set has been improved and the gap in the accuracy of answers of different lengths has become smaller. However, in TextVQA, the accuracy of the validation set decreases on answers with lengths greater than 3, and the accuracy gap between answers of different lengths becomes larger. This makes the performance improvement of the TextVQA validation set smaller than that of the ST-VQA validation set when jointly finetuning on both datasets.

\begin{figure*}[!htb]
  \centering
  \includegraphics[width=1\linewidth]{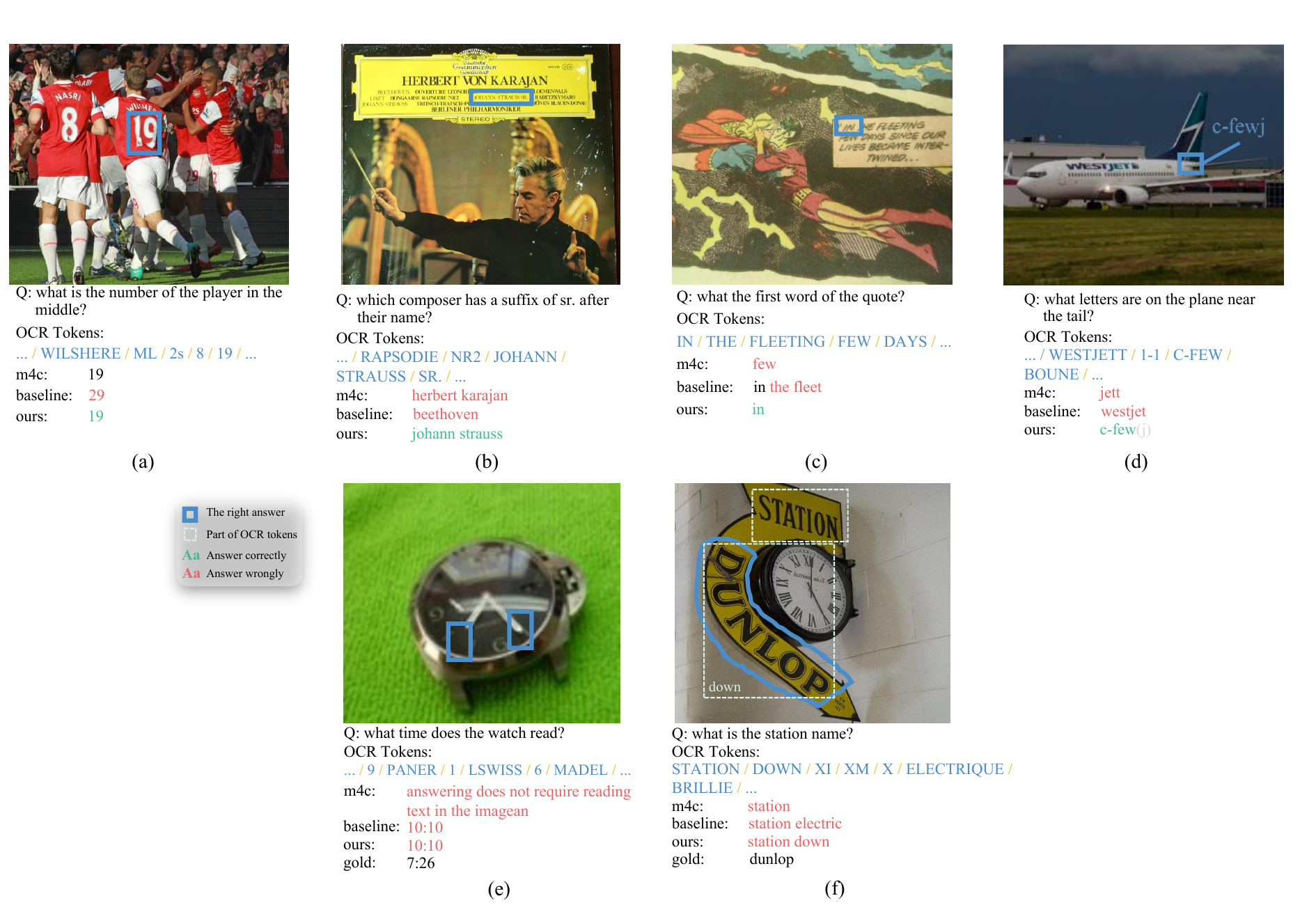}
  \caption{More qualitative examples of our model compared to the baseline model and the  M4C model. }
  \label{figure_6}
  \Description{}

\end{figure*}

\section{More Visualization Examples}

We provide more qualitative examples in Figure \ref{figure_6}  to support our claim in the paper. There are still some failure cases (Figure \ref{figure_6}) in our experiments, which are extremely challenging. 

In (a), our baseline model cannot accurately locate the correct OCR text in the image because the relative spatial relationship between each text and other text in the image is not constructed. However, our SaL model can accurately locate the correct OCR text in the middle of the image that is relevant to the question with the help of the SCP module. In (d), M4C and our baseline model tend to choose the prominent text in the image as the answer. SaL could construct the spatial position relation of OCR texts in the image and predict the right answer according to the position relation word in the question.

 All the previous models did not take into account that most of the OCR texts in the image do not have a semantic relationship, and all the OCR texts are forcibly spliced into sentences. In (b) and (c), m4c could not answer the question about a complex image with quantities of OCR texts. Our baseline model tends to predict an answer without considering the semantic relation with words in the question and image. Our TSS module separates different OCR texts and then lets the model learn whether there is a semantic relationship between different OCR texts, which greatly reduces the noise in the previous method.So even in the face of complex questions and images, SaL can answer questions very well.

However, as shown in (e), facing the problem of time, since the OCR system is difficult to identify the correct time, the input of the model does not have the correct time text information. It is difficult for our model to choose the correct answer from a huge vocabulary just based on the direction of the clock in the image. Figure \ref{figure_6}(e) is an interesting example, although our predicted answer is wrong because the OCR system misrecognized `dunlop' as `down'. But we were pleasantly surprised to find that SaL can judge that the two texts `station' and `dunlop' attached to the same material in different directions are closely related. This is not only due to the fact that SCP provides the model with the spatial relationship between different texts but also the semantic correlation between texts learned by TSS.

\end{document}